%% file: wikification2.tex
\documentclass[11pt]{article}
\usepackage{acl2015}
\usepackage{times}
\usepackage{enumitem}
\usepackage{latexsym}
\usepackage{amsmath}
\usepackage{multirow}
\usepackage{url}
\usepackage{graphicx}
\usepackage{tikz}
\usepackage{float}
\usetikzlibrary{positioning}
\usetikzlibrary{shapes,arrows,shadows}
\usepackage{subfigure}
\usepackage{amssymb}
\usepackage{array}
\usepackage{grffile}
\usepackage{epstopdf}
\usepackage{algorithmic}
\usepackage[lined,algonl,boxed]{algorithm2e}

\DeclareMathAlphabet{\mathpzc}{OT1}{pzc}{m}{it}
\usepackage{threeparttable}
\usepackage{CJKutf8}
\definecolor{Orange}{rgb}{1,0.5,0}

\newcommand{\nop}[1]{}

\title{Leveraging Deep Neural Networks and Knowledge Graphs \\for Entity Disambiguation}

\author{Hongzhao Huang\textsuperscript{1}, Larry Heck\textsuperscript{2}, Heng Ji\textsuperscript{1}\\
       \textsuperscript{1}Computer Science Department, Rensselaer Polytechnic Institute, Troy, NY 12180, USA\\
       \textsuperscript{2}Google Research, Mountain View, CA 94043, USA\\
       {\tt hongzhaohuang@gmail.com},
      {\tt larry.heck@ieee.org},\\
      {\tt jih@rpi.edu }
    }

\date{}

\begin{document}
\maketitle

\input{0abstract}

\input{1introduction}

\input{3related}

\input{4dnn}

\input{5graph}

\input{6evaluation}

\input{8conclusion}


\bibliographystyle{acl}
\bibliography{wikification2}
\end{document}

%% file: 0abstract.tex
\begin{abstract}

Entity Disambiguation aims to link mentions of ambiguous entities to a knowledge base (e.g., Wikipedia). Modeling topical coherence is crucial for this task based on the assumption that information from the same semantic context tends to belong to the same topic. This paper presents a novel deep semantic relatedness model (DSRM) based on deep neural networks (DNN) and semantic knowledge graphs (KGs) to measure entity semantic relatedness for topical coherence modeling. The DSRM is directly trained on large-scale KGs and it maps heterogeneous types of knowledge of an entity from KGs to numerical feature vectors in a latent space such that the distance between two semantically-related entities is minimized. Compared with the state-of-the-art relatedness approach proposed by \cite{milne2008b}, the DSRM obtains 19.4\% and 24.5\% reductions in entity disambiguation errors on two publicly available datasets respectively.

\end{abstract}

%% file: 1introduction.tex
\section{Introduction}

\begin{figure}[htp]
\centering
\includegraphics[width=1.0\linewidth]{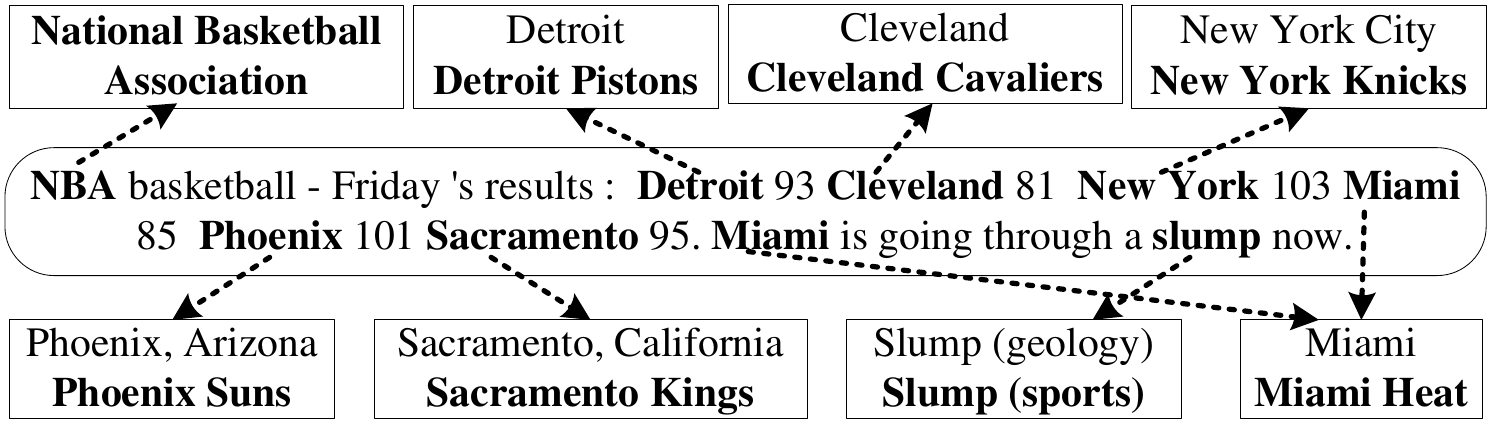}
\vspace{-1.5em}
\caption{An illustration of Entity Disambiguation task. Referent entities are marked in bold.}
\label{overall:example}
\vspace{-0.9em}
\end{figure}

Entity disambiguation is the task of linking mentions of ambiguous entities to their referent entities in a knowledge base (KB) such as Wikipedia~\cite{Mihalcea:2007}~\footnote{We consider an \emph{entity} $e$ as a page in Wikipedia or a node in knowledge graphs, and an \emph{entity mention} $m$ as an n-gram from a specific natural language text. And in this work, we focus on entity disambiguation and we assume that mentions are given as input (e.g., detected by a named entity recognition system).}. For example, the mentions (e.g., ``\emph{Detroit}'' and ``\emph{Miami}'') in Figure~\ref{overall:example} should be linked to entities related to \emph{National Basketball Association} (NBA) such as basketball teams ``\emph{Detroit Pistons}'' and ``\emph{Miami Heat}'', instead of cities ``\emph{Detroit}'' and \emph``{Miami}''.

A crucial evidence for this task is topical coherence which assumes that information from the same context tends to belong to the same topic. For instance, the text in Figure~\ref{overall:example} is on a specific topic \emph{NBA basketball}, and we can see that the mentions from this text are also linked to entities related to this topic. Modeling topical coherence normally requires to define a measure to capture semantic relatedness between candidate entities of the mentions from the same context. The standard relatedness measure widely adopted in existing disambiguation systems leveraged Wikipedia anchor links with Normalized Google Distance~\cite{milne2008b}, which can be formulated as:
\begingroup\makeatletter\def\f@size{10}\check@mathfonts
\begin{equation}
\label{semantic:relatedness:mw}
 SR^{mw}(e_i, e_j) =1 - \frac{\log{\max(|E_i|,|E_j|)}-\log{|E_i\cap E_j|}}{\log(|E|)-\log{\min(|E_i|,|E_j|)}},\nonumber
\end{equation}
\endgroup
where $|E|$ is the total number of entities in Wikipedia, and $E_i$ and $E_j$ are the set of entities that have links to $E_i$ and $E_j$, respectively. Our analysis reveals that it generates unreliable relatedness scores in many cases and tends to be biased towards popular entities. For instance, it predicts that ``\emph{NBA}'' is more semantically-related to the city ``\emph{Chicago}'' than its basketball team ``\emph{Chicago Bulls}''. 
This is because popular entities such as ``\emph{Chicago}'' tend to share more common incoming links with other entities in Wikipedia. Also, an underlying assumption of this method is that semantically-related entities must share common anchor links, which is too strong. 





\begin{figure}[htp]
\centering
\includegraphics[width=1.0\linewidth]{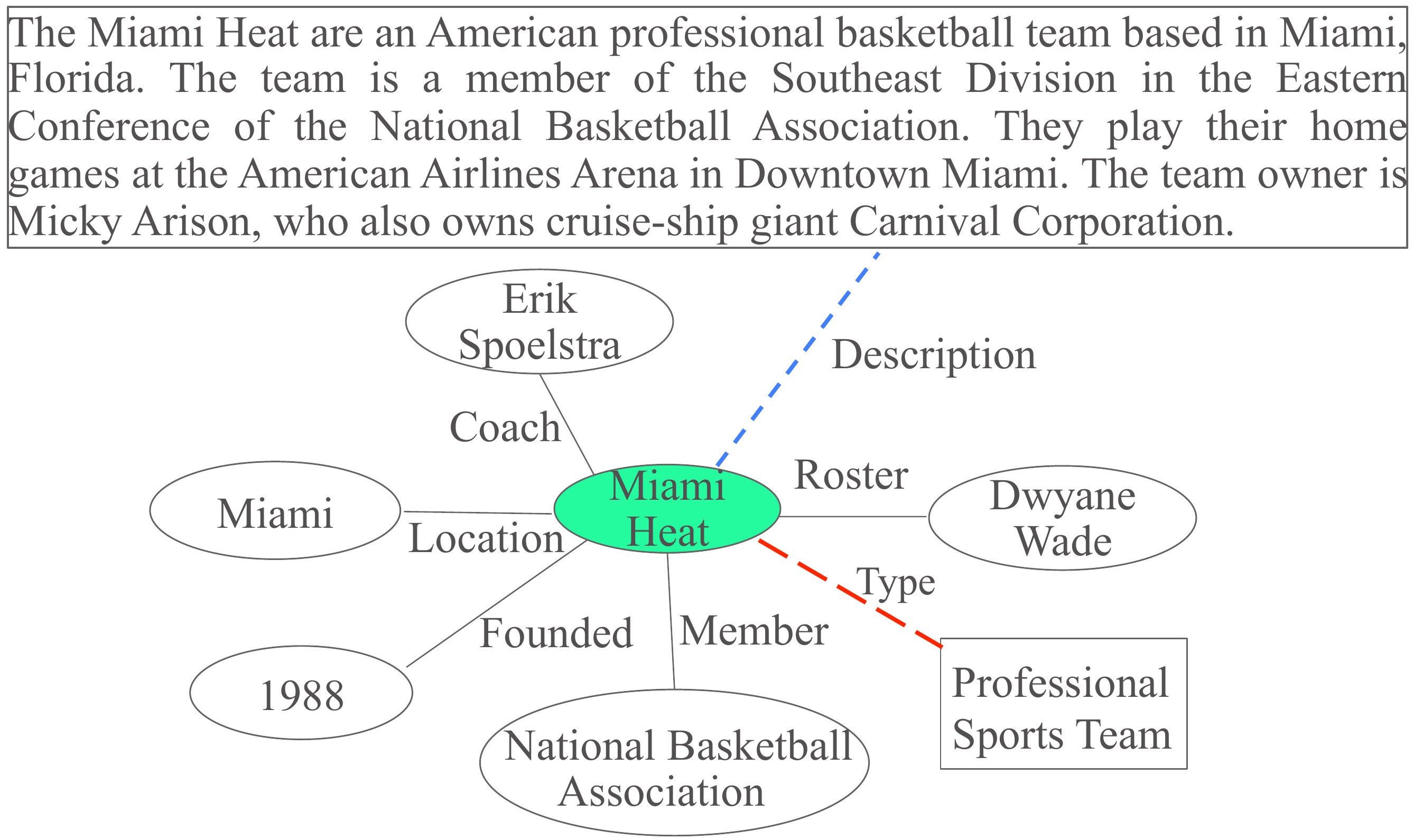}
\vspace{-1.5em}
\caption{An example of Freebase. Nodes represent entities such as ``\emph{Miami Heat}'', and edges represent semantic relations such as ``\emph{Coach}'' and ``\emph{Location}''. Each entity is also provided with \emph{textual description} and \emph{entity types}.} 
\label{semantic:graph:example}
\vspace{-0.5em}
\end{figure}

To address these limitations, we propose a novel deep semantic relatedness model (DSRM) that leverages semantic knowledge graphs (KGs) and deep neural networks (DNN). In the past decade, tremendous efforts have been made to construct many large-scale structured and linked KGs (e.g., Freebase~\footnote{https://www.freebase.com/} and DBpedia~\footnote{http://www.dbpedia.org/}), which stores a huge amount of clean and important knowledge about entities from contextual and typed information to structured facts. Each \emph{fact} is represented as a triple connecting a pair of entities by a certain relationship and of the form $\{left\: entity,\: relation,\: right\: entity\}$. An example about the entity ``\emph{Miami Heat}'' in Freebase is as shown in Figure~\ref{semantic:graph:example}. These semantic KGs are valuable resources to enhance relatedness measurement and deep understanding of entities. 


Low dimensional representations (i.e., distributed representations) of objects (e.g., words, documents, and entities) have shown remarkable success in the fields of Natural Language Processing (NLP) and information retrieval due to their ability to capture the latent semantics of objects~\cite{collobert:2011b,Huang:DSSM:2013}. Deep learning techniques have been applied sucessfully to learn distributed representations since they can extract hidden semantic features with hierarchical architectures and map objects into a latent space (e.g., \cite{Bengio:2003:NPL,collobert:2011b,Huang:DSSM:2013,Bordes:2013,Socher:2013}). Motivated by the previous work, we propose to learn latent semantic entity representations with deep learning techniques to enhance entity relatedness measurement. 
We directly encode heterogeneous types of semantic knowledge from KGs including structured knowledge (i.e., \emph{entity facts} and \emph{entity types}) and textual knowledge (i.e., \emph{entity descriptions}) into DNN. By automatically mining a large amount of training instances from KGs and Wikipedia, we then train the neural network models discriminatively in a supervised fashion such that the distances between semantically-related entities are minimized in a latent space. In this way, the neural networks can be optimized directly for the entity relatedness task and capture semantics in this dimension. 
Therefore, compared to the standard approach proposed by \cite{milne2008b}, 
our proposed DSRM is in nature a deep semantic model that can capture the latent semantics of entities. Another advantage is that it can capture more semantically-related relations between entities which do not share any common anchor links. 

The main contributions of this paper are summarized as follows:
\begin{itemize}[noitemsep,topsep=5pt,leftmargin=*]
  \item We propose a novel deep semantic relatedness model based on DNN and semantic KGs to measure entity semantic relatedness. 
   \item We explore heterogeneous types of semantic knowledge from KGs and show that semantic KGs are better resources than Wikipedia anchor links to measure entity relatedness.
   \item 
   By conducting extensive experiments on publicly available datasets from both news and tweets, we show that the proposed DSRM significantly outperforms several competitive baseline approaches regarding both relatedness measurement and entity disambiguation quality. 

\end{itemize}




%% file: 3related.tex
\section{Related Work}

Measuring semantic similarity or relatedness between words, phrases, and entities have many applications in NLP such as the entity disambiguation task studied in this work. Existing approaches mainly leveraged some classic similarity measures that do not utilize semantics or topic models and they were built on top of a thesaurus (e.g., WordNet) or Wikipedia~\cite{mchale1998comparison,Landauer1998,Strube:2006:WCS,Gabrilovich:2007,milne2008b,Ceccarelli:2013}. In contrast, we leverage both structured and contextual information from large-scale semantic KGs and deep semantic models to measure entity relatedness. 

Most existing entity disambiguation methods considered entity relatedness as a crucial evidence, from non-collective approaches that resolve one mention at each time~\cite{Mihalcea:2007,milne2008,guo2013} to collective approaches that leverage the global topical coherence for joint disambiguation~\cite{cucerzan2007,kulkarni2009,Han2011b,ratinov2011,Hoffart:2011,cassidy2012,Shen:2013,liu:2013,huang:2014}. \newcite{huang:2014} proposed a collective approach based on semi-supervised graph regularization and achieved the state-of-the-art performance for tweets. To study the impact of our proposed DSRM on entity disambiguation, we adapt their approach and develop an unsupervised approach to model topical coherence for both news and tweets.

This work is highly related to distributed representation learning of textual objects such as words, phrases, and documents with deep learning techniques (e.g., ~\cite{Bengio:2003:NPL,Mnih:2007,Hinton:2010,collobert:2011b,Bordes12jointlearning,Huang:2012,Bordes:2013,Socher:2013,NIPS2013_5021,he-EtAl:2013:Short,Huang:DSSM:2013,Shen:2014,yih2014,gao-EtAl:2014:EMNLP2014}). 
Among the above work, \newcite{Huang:DSSM:2013} is the most relevant to ours. We extend their work to large-scale semantic KGs by leveraging both structured and contextual knowledge for semantic representation learning of entities. Then we apply the approach to model topical coherence for entity disambiguation, as opposed to Web search. \newcite{he-EtAl:2013:Short} first explored deep learning techniques to measure \emph{local} context similarity for entity disambiguation. This work complements theirs since we aim to measure entity relatedness for \emph{global} topical coherence modeling.

Semantic KGs have been demonstrated to be useful resources for external knowledge mining for entity and relation extraction~\cite{Hakkani-TurHT13,HeckHT13} and coreference and entity linking~\cite{HajishirziZWZ13,Dutta:TACL522}. Some recent work learned distributed representations for entities directly from KGs for semantic parsing~\cite{Bordes12jointlearning}, link prediction~\cite{Bordes:2013,Socher:2013,wang-EtAl:2014:EMNLP20145,lin:2015}, and question answering~\cite{bordes2014,yang-EtAl:2014:EMNLP2014}. Our work leverages KGs to learn entity representations to measure entity relatedness, which is different from the above problems.

\nop{The standard approach to measure semantic relatedness between entities is based on Wikipedia anchor links with Normalized Google Distance~\cite{milne2008b,Cilibrasi:2007}, which is formulated as:
\begingroup\makeatletter\def\f@size{10}\check@mathfonts
\begin{equation}
\label{semantic:relatedness:mw}
 SR^{mw}(e_i, e_j) =1 - \frac{\log{\max(|E_i|,|E_j|)}-\log{|E_i\cap E_j|}}{\log(|E|)-\log{\min(|E_i|,|E_j|)}},\nonumber
\end{equation}
\endgroup
where $|E|$ is the total number of entities in Wikipedia, and $E_i$ and $E_j$ are the set of entities that have links to $E_i$ and $E_j$, respectively. In this paper, we propose to measure semantic relatedness from a completely different angle by exploiting deep learning techniques and semantic KGs.}

%% file: 4dnn.tex
\section{A Deep Semantic Relatedness Model (DSRM)}

In order to measure entity relatedness for topical coherence modeling, we propose to learn latent semantic entity representations that capture the latent semantics of entities. 
To learn entity representations, we directly encode various semantic knowledge from KGs into DNN.

\subsection{The DSRM Architecture}

\begin{figure}[htp]
\centering
\includegraphics[width=1.0\linewidth]{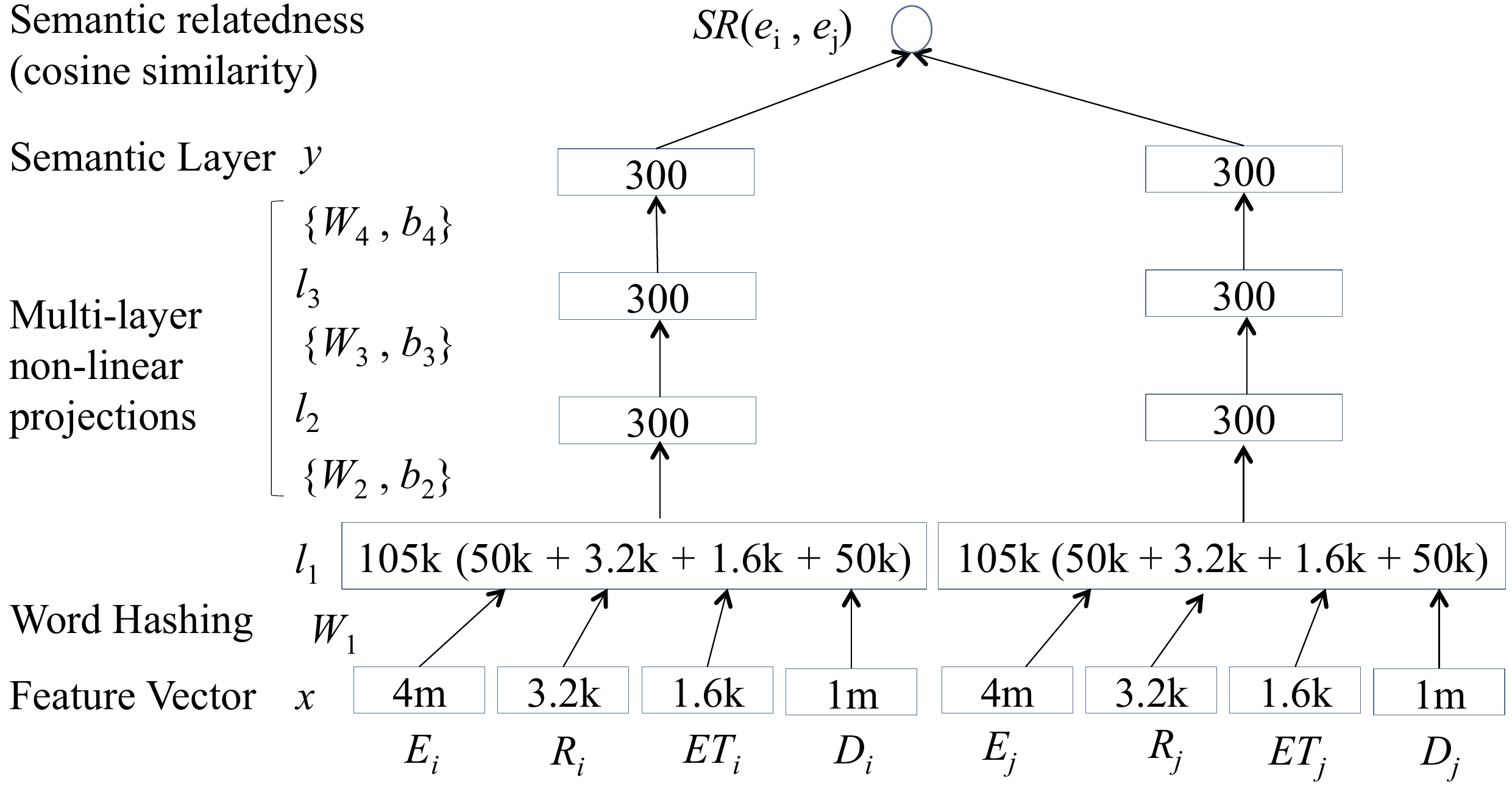}
\vspace{-2.5em}
\caption{The DSRM architecture.}
\label{DSRM:architecture}
\vspace{-0.7em}
\end{figure}

The architecture of the DSRM is shown in
Figure~\ref{DSRM:architecture}. The knowledge representations of an entity from KGs are shown in the two bottom layers (Feature Vector and Word Hashing). Following \cite{Huang:DSSM:2013}, we adopt the \emph{letter-n-gram based word hashing} technique to reduce the dimensionality of the bag-of-word term vectors. This is because the vocabulary size of the large-scale KGs is often very large (e.g., more than 4 million Wikipedia entities and 1 million bag-of-words exist in Wikipedia), which makes the ``one-hot'' vector representation very expensive. However, the word hashing techniques can dramatically reduce the vector dimensionality to a constant small size (e.g., 50k). It also can handle the out-of-vocabulary words and newly created entities. The specific approach we use is based on letter tri-grams. 
For instance, the word ``cat'' can be split into letter tri-grams (\#ca, cat, at\#) by first adding start- and
end- marks to the word (e.g., \#cat\#). We then use a vector of letter tri-grams to represent the word. In particular, we leverage four types of knowledge from KGs to represent each entity $e$, which is described in details as follows:

\nop{\begin{figure}[h]
\begin{center}
\noindent
  \includegraphics[width=3in]{fig/triletter.JPG}
  \caption{The letter tri-gram vectors of the word ``cat'' after Word Hashing.}\label{FigHotVector}
\end{center}
\vspace{-0.7em}
\end{figure}
} 

\begin{itemize}[noitemsep,topsep=5pt,leftmargin=*] 
  \item \textbf{Connected Entities $E$}: the set of connected entities of $e$. For instance, as shown in Figure~\ref{semantic:graph:example}, $E$ = \{``\emph{Erik Spoelstra}'', ``\emph{Miami}'', ``\emph{NBA}'', ``\emph{Dwyane Wade}''\} for ``\emph{Miami Heat}''. For each $e_i \in E$, we generate its surface form and represent it as bag-of-words. And then the word hashing layer transforms each word into a letter tri-gram vector. 
  \item \textbf{Relations $R$}: the set of relations that $e$ holds. For example, $R$ = \{``\emph{Coach}'', ``\emph{Location}'', ``\emph{Founded}'', ``\emph{Member}'', ``\emph{Roster}''\} for ``\emph{Miami Heat}'' in Figure~\ref{semantic:graph:example}. Each relation $r_i$ in $R$ is represented as a binary ``one-hot'' vector (e.g.., $[0,...,0,1,0,...,0]$). 
  \item \textbf{Entity Types $ET$}: the set of attached entity types for $e$. $ET$ = \{``\emph{professional sports team}''\} for ``\emph{Miami Heat}''.  We represent each entity type as a binary ``one-hot'' vector. We do not adopt word hashing to break down relations and entity types because their sizes are relatively small (i.e., 3.2k relations and 1.6k entities).
  \item \textbf{Entity Description $D$}: the textual description of an entity. The description provides a concise summary of salient information of $e$. For instance, from the description of ``\emph{Miami Heat}'', we can learn about its important information such as \emph{role}, \emph{location}, and \emph{founder}. 
The description is represented as bag-of-words, which are then transformed by the word hashing layer into letter tri-gram vectors. 
\end{itemize}

On top of the word hashing layer, we have multiple hidden layers to perform non-linear transformations, which allow the DNN to learn useful semantic features by performing back propagation with respect to an objective function designed for the entity relatedness task. Finally, we can obtain the semantic representation $y$ for $e$ from the top layer. Denoting $x$ as the input feature vector of $e$, $y$ as the output
semantic vector of $e$, $N$ as the number of layers, $l_i, i=1,...,N-1$ as the output vectors of the intermediate hidden layers, $W_i$ and $b_i$ as the weight matrix and bias term of the i-th layer respectively, we then can formally present the DSRM as:
\begin{eqnarray}
l_1 & = & W_1 x \nonumber \\
l_i & = & f(W_i l_{i-1} + b_i), \;\; i=2,...,N-1  \nonumber\\
y & = & f(W_N l_{N-1} + b_N ) \nonumber
\end{eqnarray}
where we use the $tanh$ as the activation function at the output layer and the intermediate hidden layers. Specifically, $f(x) = tanh(x) = \frac{1 - e^{-2x}}{1 + e^{-2x}}$.

After we obtain the semantic representations for entity $e_i$ and $e_j$, we use cosine similarity to measure their relatedness as $SR^{DSRM}(e_i,e_j) = \frac{{y_{e_i}^T} y_{e_j}}{||y_{e_i}|| ||y_{e_j}||}$,
\nop{\begin{equation}
SR^{DSRM}(e_i,e_j) = \cos(y_{e_i} ,y_{e_j} ) = \frac{{y_{e_i}} y_{e_j}^T}{||y_{e_i}|| ||y_{e_j}||}, \nonumber
\end{equation}} 
where $y_{e_i}$ and $y_{e_j}$ are the semantic representations of $e_i$ and $e_j$, respectively. 

\subsection{Learning the DSRM}

\textbf{Training Data Mining:} In order to train the DSRM which can capture semantics specific to the entity relatedness task, we first automatically mine training data based on KGs and Wikipedia anchor links. 
Beyond using linked entity pairs from KGs as positive training instances, we also mine more training data (especially negative instances) from Wikipedia. Suppose $t_i$ is an anchor text from a Wikipedia article, and it is linked to an entity $e_i$. And $t_j$ is an anchor text within $\delta = 150$ character window of $t_i$, and $e_j$ is its linked entity. Then we consider $\langle e_i, e_j\rangle$ as a positive training instance. 
To obtain negative training instances for $e_i$, we randomly sample $5$ other candidate entities of $t_j$ (denoted as $\hat{E_j}$), and consider each $\langle e_i, e'_j\rangle$ as a negative training instance for each $e'_j \in \hat{E_j}$. Similarly, we obtain negative training instances for $e_j$. In this way, we finally obtain about $20$ million positive training pairs and $200$ million negative training pairs. By mining the training instances automatically, we can train the DSRM in an unsupervised way and save tremendous human annotation efforts. 

\textbf{Model Training: } Following \cite{collobert:2011b,Huang:DSSM:2013,Shen:2014}, we formulate a loss function as:
\begin{equation}
\mathcal{L(\wedge)} = -\log\prod_{(e, e^+)}P(e^+| e), \nonumber
\end{equation}
where $\wedge$ denotes the set of parameters of the DSRM, and $e^+$ is a semantically-related entity of $e$. $P(e_j | e_i)$ is the posterior probability of entity $e_j$ given $e_i$ through the softmax function:
\begin{equation}
P(e_j | e_i) = \frac{\exp(\gamma SR^{DSRM}(e_i,e_j))}{\sum_{e' \in E_i}\exp(\gamma SR^{DSRM}(e_i,e'))},\nonumber
\end{equation}

where $\gamma$ is the smoothing parameter which is determined based on a held-out set, and $E_i$ is the set of related or non-related entities of $e_i$ in the training data.

To obtain the optimal solution, we need to minimize the above loss function. 
In order to avoid over-fitting, we determine model parameters with cross validation by randomly splitting the mined concept pairs into two sets: training and validation sets. We set the number of hidden layers as 2 and the number of units in each hidden layer and output layer as 300. Following \cite{Huang:DSSM:2013}, we initialize each weight matrix $W_i, i=2,...,N-1$ with a uniform distribution:
$W_i \sim \Bigg[ -\sqrt{\frac{6}{(|l_{i-1}|+|l_i|}}, \sqrt{\frac{6}{(|l_{i-1}|+|l_i|}} \quad \Bigg],$ 
where$|l|$ is the size of the vector $l$. Then we train the model with mini-batch based stochastic gradient descent~\footnote{Due to space limitation, we do not derive the derivatives of the loss function here. Readers can refer to \cite{collobert:2011b,Huang:DSSM:2013} for more details.}, and the training normally converges after 20 epochs in our experiments. We set mini-batch size of training instances as 1024. It takes roughly 72 hours to finish the model training on an NVidia Tesla K20 GPU machine.

%% file: 5graph.tex
\section{Topical Coherence Modeling with Unsupervised Graph Regularization}
\label{semantic:graph}

There exist many approaches to model topical coherence to enhance entity disambiguation. A recent approach proposed by~\cite{huang:2014} leveraged a semi-supervised graph regularization model to perform collective inference over multiple tweets and achieved the state-of-the-art performance. In this work, we adapt their approach and develop a completely unsupervised framework which is more suitable for the entity disambiguation task. This is because it is challenging to obtain manually labeled seeds for new and unseen data (e.g., a news document).


\subsection{Relational Graph Construction}

\begin{figure}[htp]
\centering
\includegraphics[width=0.85\linewidth]{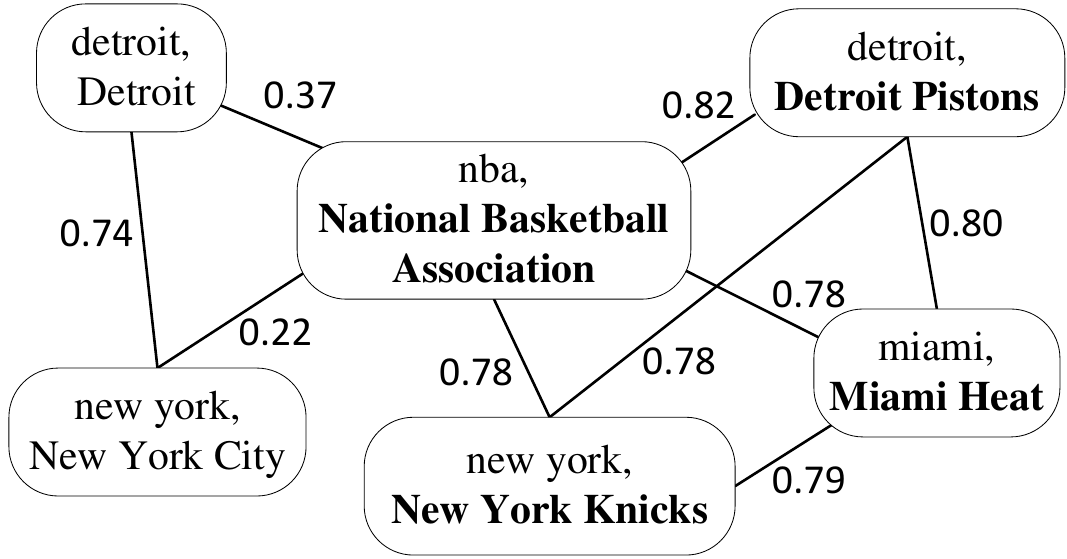}
\vspace{-0.7em}
\caption{A portion of the relational graph constructed for the example in Figure \ref{overall:example}. The entities marked in bold are the referent entities for the mentions in the same node.}
\vspace{-0.7em}
\label{relational:graph}
\end{figure}

We first construct a relational graph $G=\langle V, E\rangle$~\footnote{In this work, we choose not to incorporate the set of local features adopted in \cite{huang:2014} since they are mainly designed for mention detection instead of disambiguation.}, where $V$ is a set of nodes and $E$ is a set of edges. Each node $v_i=\langle m_i, e_i\rangle$ contains a pair of mention $m_i$ and its entity candidate $e_i$.  Each node $v_i$ is also associated with a ranking score $r_i$, indicating the probability of $e_i$ being the referent entity of $m_i$. 

A weighted edge is added between two nodes $v_i$ and $v_j$ if and only if (i) $m_i$ and $m_j$ are relevant, (ii) $e_i$ and $e_j$ are semantically-related, (iii) $v_j$ is one of the $k$-th nearest neighboring nodes of $v_i$~\footnote{We set k as 20, which is obtained from a development set.}. Let $W$ be the weight matrix of the relational graph $G$. Then if $v_i$ and $v_j$ satisfy the above three conditions, the weight of their connecting edge is computed as: $W_{ij} = SR(e_i, e_j)$, where $SR(e_i, e_j)$ is a relatedness measure. Otherwise, $W_{ij}$ is set as 0. To determine whether $m_i$ and $m_j$ are relevant or not in tweets, we follow~\cite{huang:2014} to check their connectivity in a network constructed from social relations such as \emph{authorship} and \emph{\#hashtag}. For news, we only model topical coherence within one single document ($m_i$ and $m_j$ are relevant if they are from the same document). This is because each news document normally has relative rich contextual information, and we do not have adequate social relations such as authorship relation in the dataset. 
An example relational graph is shown in Figure~\ref{relational:graph}.

\nop{Let $W$ be the weight matrix of the relational graph $G$. Then we compute a k nearest neighbor (kNN) graph, where each node is connected to its k (i.e., k = 20) nearest neighboring nodes. It can be represented as:
\[ W_{ij}=\left\{
	\begin{array}{l l}
        rel(m_i, m_j) * SR(e_i, e_j) & \quad j\in kNN(i)\\
        0 & \quad \text{Otherwise}
	\end{array} \right. \]
where $SR(e_i, e_j)$ is a relatedness measure. }

\subsection{Ranking Score Initialization and Automatic Labeled Seed Mining}

We initialize the ranking score of each node based on a sub-system of AIDA~\cite{Hoffart:2011}, which relies on the linear combination of prior popularity and context similarity. The prior popularity 
is measured based on the frequency of Wikipedia anchor links. The context similarity proposed in AIDA is computed based on the extracted keyphrases (e.g., Wikipedia anchor texts) of an entity and all of their partial matches in the text of a mention.

We also adopt two heuristics to mine a set of labeled seed nodes for the graph regularization model: (i) If a node $v$ contains unambiguous mention, then $v$ is selected as a seed node and it has an initial ranking score 1.0. (ii) For a mention $m$ with the top ranked candidate entity by prior popularity as $e$, if the prior popularity $p(e|m)$ of $e$ satisfies $p(e|m) \geq 0.95$ and $e$ is also the top ranked entity by context similarity, then all nodes related to $m$ are selected as labeled seeds. The node $v = \langle m, e \rangle$ will be assigned a ranking score 1.0, and other nodes will be assigned a ranking score 0. During the graph regularization process, the ranking scores of these labeled seed nodes will remain unchanged.

\subsection{Graph Regularization}

We then utilize the graph regularization model to refine the ranking scores of unlabeled nodes simultaneously. Denoting the first $l$ nodes as seed nodes with initial ranking scores as $R_l$ ($l=0$ is possible in some cases if our approach fails to find some labeled seeds), the remaining $u$ nodes ($u=n-l$) are initialized with ranking scores $R_u^{0}$, and $W$ as the weight matrix of the relational graph $G$. Then the graph regularization framework~\cite{ZhuGL03,huang:2014} can be formulated as $\mathcal{F(R)} =  \mu\sum_{i={l+1}}^{n}(r_{i}-r^{0}_{i})^{2} + \frac{1}{2}\sum_{i, j}W_{ij}(r_i - r_j)^{2}$, where $\mu$ is a regularization parameter that controls the trade-off between initial rankings and smoothness over the graph structure. This loss function aims to ensure the constraint that two strongly connected nodes should have similar ranking scores. 
There exist both closed-form and iterative solutions for the above optimization problem since $\mathcal{F(R)}$ is convex~\footnote{Readers can refer to \cite{ZhuGL03,huang:2014} for the derivation of the closed-form and iterative solutions.}~\cite{ZhuGL03,huang:2014}.

%% file: 6evaluation.tex
\section{Experiments}

In this section, we evaluate the performance of various semantic relatedness methods and their impact on the entity disambiguation task. 

\nop{
\begin{table*}[h!t]
\centering
\begin{threeparttable}
\footnotesize
\begin{tabular}{|p{1cm}|p{14cm}|}
 \hline
 \textbf{Methods} & Descriptions\\
 \hline
 \textbf{Kul\_sp} & This is a collective inference approach relying on integer linear programs~\cite{kulkarni2009}. \\
 \hline
 \textbf{Shirak} & This approach utilizes a probabilistic taxonomy with a naive Bayes probabilistic model to perform disambiguation~\cite{Shirakawa:2011}.\\
 \hline
 \textbf{AIDA} & This is a graph-based collective inference approach proposed in \cite{Hoffart:2011}, which aims to find a dense subgraph for optimal joint disambiguation.\\
 \hline
 \textbf{TagMe} & This approach determines the referent entity by computing the sum of weighted average semantic relatedness scores between entities~\cite{ferragina:2010}.\\
 \hline
 \textbf{Meij} & The best system proposed by \cite{Meij:2012}, which is a supervised approach using random forest model and various local features.\\
 \hline
\end{tabular}
\end{threeparttable}
\vspace{-0.2cm}
\caption{Description of baseline methods.} \label{table:methods}
\vspace{-0.7em}
\end{table*}
}

\input{6.1data}

\input{6.2setting}
\input{6.2relatedness}

\input{6.3overall}

\input{6.4discussions}

%% file: 6.1data.tex
\subsection{Data and Scoring Metric}

For our experiments we use Wikipedia 
as our knowledge base, which originally contains $30$ million entities. To reduce noise, we remove the entities which have fewer than 5 incoming anchor links and obtain $4$ millions entities. And we use a portion of Freebase limited to Wikipedia entities as the semantic KG with detailed statistics shown in Table~\ref{TableStats}. To evaluate the quality of entity relatedness, we use a benchmark test set created by~\cite{Ceccarelli:2013} from CoNLL 2003 data. It includes $3,314$ entities as testing queries and each query has 91 candidate entities in average to measure relatedness. 
After obtaining the ranked orders of candidate entities for these queries, we compute the nDCG~\cite{Jarvelin2002} and mean average precision (MAP)~\cite{Manning:2008:IIR:1394399} scores to evaluate the relatedness measurement quality.

To evaluate disambiguation performance, we use two public test sets composed of both news documents and tweets: (i) AIDA~\cite{Hoffart:2011} is a news dataset based on CoNLL 2003 data. It includes 131 documents and 4,485 non-NIL mentions. 
(ii) A tweet set released by \cite{Meij:2012}, which includes $502$ tweets and $812$ non-NIL mentions. 
We follow the previous work \cite{cucerzan2007} and leverage Wikipedia anchor links to construct a mention-entity dictionary for candidate generation. And for efficiency, we only consider the top 30 ranked entities by prior popularity in our systems. We compute both standard micro (aggregates over all mentions) and macro (aggregates over all documents) precision scores over the top ranked candidate entities for disambiguation performance evaluation. 

\begin{table}[t]
\centering
\small
\begin{threeparttable}
\begin{tabular}{|l|c|}
\hline
Knowledge Graph Element	& Size \\ \hline
\# Entities      	& 4.12m\\ \hline 
\# Relations		& 3.17k\\ \hline
\# Entity Types		& 1.57k\\ \hline
\end{tabular}
\end{threeparttable}
\vspace{-0.7em}
\caption{Statistics of Freebase KG.} 
\label{TableStats}
\vspace{-0.9em}
\end{table}


%% file: 6.2setting.tex
\subsection{Experimental Settings}

\begin{table*}[h!t]
\centering
\begin{threeparttable}
\footnotesize
\begin{tabular}{|p{1.5cm}|p{13.5cm}|}
 \hline
 \textbf{Methods} & Descriptions\\
 \hline
 M\&W & The Wikipedia anchor link-based method proposed by \cite{milne2008b}.\\
 \hline
 DSRM$_1$ & Our proposed DSRM based on connected entities.\\
 \hline
 DSRM$_{12}$ & DSRM$_1$ + relations.\\
 \hline
 DSRM$_{123}$ & DSRM$_{12}$ + entity types.\\
 \hline
 DSRM$_{1234}$ & DSRM$_{123}$ + entity descriptions. \\
 \hline
\end{tabular}
\end{threeparttable}
\vspace{-0.2cm}
\caption{Description of semantic relatedness methods.} \label{table:semantic:methods}
\vspace{-0.7em}
\end{table*}

The semantic relatedness measures we study in this work are summarized in Table~\ref{table:semantic:methods}. We combine these measures with the unsupervised graph regularization model (GraphRegu) and develop an unsupervised collective inference framework for entity disambiguation. We compare our methods with several state-of-the-art entity disambiguation approaches as follows (The first three methods were developed for news, and \emph{TagMe} and \emph{Meij} were proposed for tweets):

\begin{itemize}[noitemsep,topsep=5pt,leftmargin=*]
   \item \textbf{Kul\_sp}: This is a collective approach with integer linear programs~\cite{kulkarni2009}.
   \item \textbf{Shirak}: This approach utilizes a probabilistic taxonomy with the Naive Bayes model~\cite{Shirakawa:2011}.
   \item \textbf{AIDA}: This is a graph-based collective approach which finds a dense subgraph for joint disambiguation~\cite{Hoffart:2011}.
   \item \textbf{TagMe}: This approach determines the referent entity by computing the sum of weighted average semantic relatedness scores between entities~\cite{ferragina:2010}.
  \item \textbf{Meij}: A supervised approach based on the random forest model~\cite{Meij:2012}.
  \item \textbf{GraphRegu + a relatedness measure}: This is the collective inference framework we develop by combining the GraphRegu with a relatedness measure for the relational graph construction. Specially note that \emph{GraphRegu + M\&W} can be considered as a baseline approach we adapt from \cite{huang:2014}.
\end{itemize}

Next, we study the relatedness measurement quality of various relatedness methods (Section~\ref{relatedness:eval}), and the impact of relatedness methods on entity disambiguation (Section~\ref{disambiguation}).

%% file: 6.2relatedness.tex
\subsection{Quality of Semantic Relatedness Measurement}
\label{relatedness:eval}
The overall performance of various relatedness methods are shown in Table~\ref{overall:relatedness}. We can see that the DSRM significantly outperforms the standard relatedness method M\&W ($p\leq 0.05$, according to the Wilcoxon Matched-Pairs Signed-Ranks Test), indicating that deep semantic models based on semantic KGs are more effective for relatedness measurement. 
As we incorporate more types of knowledge into the DSRM, it achieves better relatedness quality, showing that the four types of semantic knowledge complement each other.  

To study the main differences between M\&W and the DSRM, we also show some examples of relatedness scores in Table~\ref{sample1} and \ref{sample3}. 
From both tables, we can see that M\&W predicts that ``\emph{NBA}'' and ``\emph{NFL}'' are more semantically-related to cities/states than their professional teams. However, the DSRM produces more reasonable scores to indicate that these sports teams are highly semantically-related to their association. 
We can also see that M\&W tends to generate high relatedness scores for popular entities (e.g., ``\emph{New York City}'' and ``\emph{Philadelphia}''), but the DSRM does not have such a bias. 

\begin{table}[t]
\small
\centering
\begin{threeparttable}
\begin{tabular}{|p{1.22cm}|p{1.1cm}|p{1.1cm}|p{1.3cm}|p{0.6cm}|}
\hline
Methods & nDCG@1 & nDCG@5 & nDCG@10 & MAP \\ 
\hline
M\&W & 0.54 & 0.52 & 0.55 & 0.48 \\ 
\hline
DSRM$_1$ & 0.68 & 0.61 & 0.62 & 0.56\\
\hline 
DSRM$_{12}$ & 0.72 & 0.64 & 0.65 & 0.59 \\
\hline
DSRM$_{123}$ & 0.74 & 0.65 & 0.66 & 0.61 \\
\hline
DSRM$_{1234}$ & \textbf{0.81} & \textbf{0.73} & \textbf{0.74} & \textbf{0.68} \\
\hline
\end{tabular}
\end{threeparttable}
\vspace{-0.7em}
\caption{Overall performance of entity semantic relatedness methods.} \label{overall:relatedness}
\vspace{-0.9em}
\end{table}

\begin{table}[t]
\small
\centering
\begin{threeparttable}
\begin{tabular}{|l|c|c|}
\hline
Methods       & M\&W & DSRM$_{1234}$\\ 
\hline
New York City & 0.90 & 0.22 \\ 
\hline 
New York Knicks	& 0.79 & \textbf{0.79}\\
\hline
Boston & 0.75 & 0.27\\
\hline
Boston Celtics & 0.58 & \textbf{0.77}\\
\hline
Dallas & 0.69 &  0.35\\
\hline 
Dallas Mavericks & 0.52 & \textbf{0.74}\\
\hline
Philadelphia & 0.81 & 0.27\\
\hline
Philadelphia 76ers & 0.58 & \textbf{0.85}\\
\hline
\end{tabular}
\end{threeparttable}
\vspace{-0.7em}
\caption{Examples of relatedness scores between a sample of entities and the entity ``\textbf{NBA}''.}
\label{sample1}
\end{table}

\begin{table}[t]
\small
\centering
\begin{threeparttable}
\begin{tabular}{|l|c|c|}
\hline
Methods       & M\&W & DSRM$_{1234}$\\ 
\hline
New York City & 0.89 & 0.09 \\ 
\hline 
New York Jets	& 0.92 & \textbf{0.63}\\
\hline
Boston & 0.92 & 0.19\\
\hline
Boston Bruins & 0.62 & \textbf{0.38}\\
\hline
Dallas & 0.87 &0.34\\
\hline
Dallas Cowboys & 0.72 & \textbf{0.68}\\
\hline
Philadelphia & 0.93 & 0.19\\
\hline
Philadelphia Eagles & 0.79 & \textbf{0.65}\\
\hline
\end{tabular}
\end{threeparttable}
\caption{Examples of relatedness scores between a sample of entities and the entity ``\textbf{National Football League}''.}
\label{sample3}
\vspace{-0.9em}
\end{table}

%% file: 6.3overall.tex
\subsection{Impact on Entity Disambiguation}
\label{disambiguation}
\begin{table*}[h!t]
\centering
\begin{threeparttable}
\footnotesize
\begin{tabular}{{l}*{4}{c}|*{4}{c}}
 \hline
 &\multicolumn{4}{c|}{\textbf{Baseline Approaches}}& \multicolumn{4}{c}{\textbf{Our Methods}}\\
 \hline
 & Kul\_sp & Shirak & AIDA & \multicolumn{1}{|c|}{GraphRegu + } & \multicolumn{4}{c}{GraphRegu + }\\
 & \multicolumn{3}{c}{}      & \multicolumn{1}{|c|}{M\&W} & DSRM$_1$ & DSRM$_{12}$ & DSRM$_{123}$ & DSRM$_{1234}$\\
\hline
micro P@1.0 & 72.87 & 81.40 & 82.29 & \multicolumn{1}{|c|}{\textbf{82.23}} & 84.17 & 85.33 & 84.91 & \textbf{86.58} \\
macro P@1.0 & 76.74 & 83.57 & 82.02 & \multicolumn{1}{|c|}{\textbf{81.10}} & 83.30 & 83.94 & 83.56 & \textbf{85.47} \\
 
\end{tabular}
\end{threeparttable}
\vspace{-0.7em}
\caption{Overall disambiguation performance (\%) on AIDA dataset.} \label{results:aida}
\end{table*}

\begin{table*}[h!t]
\centering
\begin{threeparttable}
\footnotesize
\begin{tabular}{{l}*{3}{c}|*{4}{c}}
 \hline
 &\multicolumn{3}{c|}{\textbf{Baseline Approaches}}& \multicolumn{4}{c}{\textbf{Our Methods}}\\
 \hline
 & TagMe & Meij & \multicolumn{1}{|c|}{GraphRegu + } & \multicolumn{4}{c}{GraphRegu + }\\

 & &                    & \multicolumn{1}{|c|}{M\&W}  & DSRM$_1$ & DSRM$_{12}$ & DSRM$_{123}$ & DSRM$_{1234}$\\
 \hline
micro P@1.0 & 61.03 & 68.33 & \multicolumn{1}{|c|}{\textbf{65.13}}  & 69.19 & 70.22 & 71.50 & \textbf{71.89} \\
macro P@1.0 & 60.46 & 69.19 & \multicolumn{1}{|c|}{\textbf{66.20}}  & 69.04 & 69.61 & 70.92 & \textbf{71.72} \\
 
\end{tabular}
\end{threeparttable}
\vspace{-0.7em}
\caption{Overall disambiguation performance (\%) on tweet set.} \label{results:tweet}
\vspace{-0.7em}
\end{table*}

The regularization parameter $\mu$ of the GraphRegu is set as $0.8$ for both datasets, obtained from a held-out set from CoNLL 2003 data. The overall disambiguation performance is shown in Table~\ref{results:aida} and \ref{results:tweet} for the AIDA dataset and the tweet set, respectively. Compared with other strong baseline approaches, our developed unsupervised approach \emph{GraphRegu + M\&W} adapted from \cite{huang:2014} achieves very competitive performance for both datasets, illustrating that the GraphRegu is effective to model topical coherence for entity disambiguation. 



Our best system based on the DSRM with all four types of knowledge (denoted as DSRM$_{1234}$) significantly outperforms various strong baseline competitors for both datasets (all with $p\leq 0.05$). Specially compared with the standard method M\&W, 
DSRM$_{1234}$ achieves 24.5\% and 19.4\% relative reductions in disambiguation errors for news and tweets, respectively. 
For instance,  GraphRegu + M\&W fails to disambiguate the mention ``\emph{Middlesbrough}'' to the football club ``\emph{Middlesbrough F.C.}'' in the text ``\emph{Lee Bowyer was expected to play against Middlesbrough on Saturday.}''. This is because M\&W generates the same semantic relatedness score ($0.39$) between $\langle$``{Middlesbrough F.C.'', ``Lee Bowyer''$\rangle$ and $\langle$``Middlesbrough'' and ``Lee Bowyer''$\rangle$. However, DSRM$_{1234}$ computes the relatedness score for the former pair as $0.68$, much higher than the score $0.33$ of the latter one, thus GraphRegu + DSRM$_{1234}$ correctly disambiguates the mention.

%% file: 6.4discussions.tex
\subsection{Discussion}

In this subsection, we aim to answer two questions: (i) Are semantic KGs better resources than Wikipedia anchor links for relatedness measurement?  (ii) Is the DNN a better choice than Normalized Google Distance (NGD)~\cite{Cilibrasi:2007} and Vector Space Model (VSP) \cite{Salton:1975:VSM} for relatedness measurement?

In order to answer these two questions, we directly apply NGD and VSP with the tf-idf representations on the same KG that we use to learn the DSRM. Then we combine them with the graph regularization model and study their impact on entity disambiguation. Table~\ref{impact:relatedness} and \ref{impact:disambiguation} show the relatedness quality and disambiguation performance, respectively. As shown in the first three rows of both tables, we can clearly see that NGD and VSP based on KGs significantly outperform their variants with Wikipedia anchor links ($p\leq 0.05$), which confirms that semantic KGs are better resources than the Wikipedia anchor links for relatedness measurement. This is because KGs contain cleaner semantic knowledge about entities than Wikipedia anchor links. For instance, ``\emph{Apple Inc.}'' and ``\emph{Barack Obama}'' share many noisy incoming links (e.g., ``\emph{Austin, Texas}'' and ``\emph{2010s}'') that are not helpful to capture their relatedness.

From the last three rows of Table~\ref{impact:relatedness} and \ref{impact:disambiguation}, we can see that the DSRM based on DNN significantly outperform NGD and VSP for both relatedness measurement and entity disambiguation ($p\leq 0.05$), illustrating that the DNN are indeed more effective to measure entity relatedness. By extracting useful semantic features layer by layer with nonlinear functions and transforming sparse binary “one-hot” vectors into low-dimensional feature vectors in a latent space, the DNN has better ability to represent entities semantically.

\begin{table}[t]
\small
\centering
\begin{threeparttable}
\begin{tabular}{|p{1.22cm}|p{1.1cm}|p{1.1cm}|p{1.3cm}|p{0.6cm}|}
\hline
Methods & nDCG@1 & nDCG@5 & nDCG@10 & MAP \\ 
\hline
M\&W & 0.54 & 0.52 & 0.55 & 0.48 \\ 
\hline 
M\&W$_{1234}$ & 0.69 & 0.58 & 0.58 & 0.51\\
\hline
VSP$_{1234}$ & 0.68 & 0.58 & 0.58 & 0.52\\
\hline 
DSRM$_{1234}$ & \textbf{0.81} & \textbf{0.73} & \textbf{0.74} & \textbf{0.68} \\
\hline
\end{tabular}
\end{threeparttable}
\vspace{-0.7em}
\caption{Impact of semantic KGs and DNN on entity semantic relatedness.} \label{impact:relatedness}
\end{table}

\begin{table}[t]
\small
\centering
\begin{threeparttable}
\begin{tabular}{|l|c c|c c|}
\hline
Methods &\multicolumn{2}{c|}{AIDA dataset}&\multicolumn{2}{c|}{Tweet set}\\
        & micro & macro & macro & macro \\ 
        & P@1.0 & P@1.0 & P@1.0 & P@1.0 \\
\hline
M\&W           & 82.23 & 81.10 & 65.13 & 66.20\\ 
\hline 
M\&W$_{1234}$  & 84.57  & 83.81 & 68.21 & 69.17\\
\hline
VSP$_{1234}$  & 84.83 & 83.49 & 69.36 & 70.20\\ 
\hline 
DSRM$_{1234}$ & \textbf{86.58} & \textbf{85.47} & \textbf{71.89} & \textbf{71.72}\\
\hline
\end{tabular}
\end{threeparttable}
\vspace{-0.7em}
\caption{Impact of semantic KGs and DNN on entity disambiguation.} \label{impact:disambiguation}
\vspace{-0.9em}
\end{table}

%% file: 8conclusion.tex
\section{Conclusions and Future Work}
\label{section_conclusion}
We have introduced a deep semantic relatedness model based on DNN and semantic KGs for entity relatedness measurement. By encoding various semantic knowledge from KGs into DNN with multi-layer non-linear transformations, the DSRM can extract useful semantic features to represent entities. By developing an unsupervised graph regularization approach to model topical coherence with the proposed DSRM, we have achieved significant better performance than state-of-the-art entity disambiguation approaches. This work sheds light on exploring and modeling large-scale KGs with deep learning techniques for entity disambiguation and other NLP tasks. Our future work includes direct encoding of semantic paths from KGs into neural networks. 
We also plan to design a joint model for entity disambiguation and entity relatedness measurement, which allows mutual improvement of both tasks and generates dynamic and context-aware entity relatedness scores.

